\def\year{2019}\relax
\documentclass[letterpaper]{article} 
\usepackage{aaai19}  
\usepackage{times}  
\usepackage{helvet}  
\usepackage{courier}  
\usepackage{url}  
\usepackage{graphicx}  
\usepackage{amssymb,amsmath,amsthm,xspace}
\usepackage{multirow}
\usepackage{xspace}
\usepackage{tabularx}
\usepackage{xcolor}
\usepackage[toc,page]{appendix}
\usepackage{titlesec}
\usepackage{float}

\titlespacing*{\paragraph}{0pt}{1.5ex }{1em}

\newcommand{\identity}{\texttt{IDENTITY}\xspace}
\nocopyright

\theoremstyle{definition}
\newtheorem{definition}{Definition}

\begin{document}
 
%
\title{Counterfactual Fairness in Text Classification through Robustness}

 \author{Sahaj Garg,\textsuperscript{1}\textsuperscript{*}
Vincent Perot,\textsuperscript{2}
Nicole Limtiaco,\textsuperscript{2}
Ankur Taly,\textsuperscript{3}
Ed H. Chi,\textsuperscript{3}
Alex Beutel\textsuperscript{2}
\\
\textsuperscript{1}{Stanford University, Stanford, CA}\\
\textsuperscript{2}{Google AI, New York, NY}\\
\textsuperscript{3}{Google AI, Mountain View, CA}\\
\textsuperscript{*}{Work done while the author was an intern at Google.}\\
sahajg@cs.stanford.edu, \{vperot, nlimtiaco, ataly, edchi, alexbeutel\}@google.com}

\maketitle

\begin{abstract}
In this paper, we study counterfactual fairness in text classification, which asks the question: \emph{How would the prediction change if the sensitive attribute referenced in the example were different?} Toxicity classifiers demonstrate a counterfactual fairness issue by predicting that ``Some people are gay'' is toxic while ``Some people are straight'' is nontoxic. We offer a metric, counterfactual token fairness (CTF), for measuring this particular form of fairness in text classifiers, and describe its relationship with group fairness. Further, we offer three approaches, blindness, counterfactual augmentation, and counterfactual logit pairing (CLP), for optimizing counterfactual token fairness during training, bridging the robustness and fairness literature. Empirically, we find that blindness and CLP address counterfactual token fairness. The methods do not harm classifier performance, and have varying tradeoffs with group fairness. These approaches, both for measurement and optimization, provide a new path forward for addressing fairness concerns in text classification.
\end{abstract}

\section{Introduction}

Consider a model that determines whether an Internet forum comment is toxic. We would like to improve the model's fairness with respect to the {content} of the input text, which may reference sensitive identity attributes, such as sexual orientation, race, or religion. 
In practice, \citeauthor{Dixon18} showed that a toxicity model had a high false positive rate on examples that included identity tokens such as ``gay,'' because such tokens occur relatively frequently in examples labeled toxic in the training set. 

A related issue to users arises when nearly identical sentences referencing different identity groups receive  different predictions. For instance, a baseline toxicity model predicts that ``Some people are gay'' is 98\% likely to be toxic and ``Some people are straight'' is only 2\% likely to be toxic. In this work, we seek to specifically address this fairness issue for text classification.

Given an example, we ask a \emph{counterfactual} question: \emph{How would the prediction change if the sensitive attribute referenced in the example were different?} 
If the prediction score changes with respect to a sensitive attribute, we consider this an indicator of a potential problem.
In contrast to group-based notions of fairness (e.g., demographic parity, equality of odds), which seek to statistically equalize the model's behavior for entire sensitive groups, counterfactual fairness requires equal model behavior on individual counterfactual pairs; see~\cite{Kusner17,Wachter17}. 

To assess counterfactual fairness, we consider perturbations obtained by substituting tokens associated with identity groups. For instance, substituting ``gay'' with ``straight,'' or ``Asian'' with ``American.'' Based on these generated counterfactuals, we can define a fairness metric, which we call \emph{counterfactual token fairness} (CTF). While this is more limited than general counterfactual fairness, we believe it captures one of the most salient issues in text classification and is a starting point for more general counterfactual fairness metrics for text.

Deciding when counterfactual pairs should have the same prediction raises difficult ethical and philosophical questions. Many logical counterfactuals generated by token substitution may not require identical output. We call these \emph{asymmetric counterfactuals}. In toxicity classification, such situations could arise when the comment references stereotypes associated with one group but not another, or when comments attack a particularly vulnerable group. Asymmetric counterfactuals suggest that practitioners should be careful in both training and evaluation of counterfactual fairness. We discuss proposals for addressing this in the case of toxicity classification in the experiments section. 

To satisfy counterfactual token fairness, we borrow techniques from the robustness literature. We propose a general training scheme for achieving arbitrary counterfactual fairness by extending logit pairing~\cite{ALP} to penalize differences in the model's outputs for counterfactual pairs. We compare this method to simply augmenting the training set with counterfactual examples, and to blindness, which replaces all sensitive tokens with a special token. 

One issue is that the aforementioned methods may only achieve fairness with respect to identity tokens considered by counterfactuals during training. To address this, we evaluate the generalization of the methods on a held-out set of identity tokens.
Another concern when optimizing for counterfactual fairness is potential trade-offs with other desirable properties of a classifier, including overall accuracy and group fairness. In practice, we do not find significant harms with respect to accuracy, and varying effects on group fairness in the form of tradeoffs between true negatives and true positives. 

We make the following contributions:
\begin{itemize}
    \item \textbf{Metric:} We provide a tractable metric, \emph{counterfactual token fairness}, for measuring counterfactual fairness in text classification.
    \item \textbf{Methods:} We study three methods for addressing counterfactual token fairness: (A) blindness, (B) counterfactual augmentation, and (B) counterfactual logit pairing, bridging research from the robustness and fairness domains.
    \item \textbf{Empirical Evaluation:} We evaluate empirical performance and tradeoffs of counterfactual token fairness, group fairness, and accuracy across these approaches.
\end{itemize}

\section{Related Work} \label{related}

\paragraph{ML Fairness} Significant work in the ML fairness literature has been devoted to measuring fairness. Our work is most closely related to counterfactual fairness in causal inference \cite{Kusner17,Kilbertus17}, where fairness is evaluated by applying counterfactual interventions over a causal graph. Our definition of counterfactual token fairness implicitly defines a simple causal model for text generation. \citeauthor{Kusner17} also draw the connection between counterfactual fairness and individual fairness, which requires similar predictions for similar inputs via a Lipschitz constraint  \cite{Dwork11}.  

Relatively more study has been devoted to group fairness metrics, which evaluate observational criteria, or statistical relationships between the data, group membership, the label, and the model's prediction. Such metrics include demographic parity and equality of odds \cite{Hardt16}. \citeauthor{Hardt16} demonstrate that observational criteria are insufficient to distinguish between some seemingly reasonable uses of identity and other unreasonable ones. This is because observational criteria cannot incorporate any external understanding about what is causally acceptable in making predictions. The limitations of observational criteria can be addressed by counterfactual or individual fairness, see \cite{Kusner17,Kilbertus17,Dwork11}. By extending these definitions to path-specific counterfactual fairness, it is possible to specify which uses of identity are acceptable \cite{Chiappa18}.

Social science literature on fairness raises arguments for counterfactual reasoning as well as potential limitations. One concern is about the ability to reasonably intervene on identity of an individual. Given that most social scientists agree that race is socially constructed, it may be unreasonable to attempt to modify race and all its associated factors \cite{KH19}. These limitations, among others, are reflected in debate surrounding the use of counterfactuals over race in epidemiological studies \cite{Krieger14,VWT14}. We note that our work deals with well-defined interventions on content by only manipulating identity tokens in text, rather than the actual identities of individuals, which differentiates it from the work above. 

ML fairness literature has also focused on debiasing methods to address these gaps. Many methods have been proposed to address group fairness issues, such as re-calibrating score functions \cite{Hardt16}, adversarially learning fair representations \cite{Zemel13,VFAE,Beutel17}, data rebalancing \cite{Dixon18}, and data augmentation using swaps of gender terms \cite{Park18}. For natural language problems, \citeauthor{Pryzant18} learn a lexicon that is uncorrelated to a set of confounding variables. Debiasing methods for counterfactual or individual fairness have been studied less for neural network models. The methods in \cite{Kusner17,Kilbertus17} are effective for causal graphs, but most machine learning problems will not fit this mold. To address individual fairness, \cite{Dwork11} applies constraint based optimization over a linear program, but it is difficult to define valid distance metrics or apply the optimization to arbitrary neural networks used in natural language processing. 

\paragraph{Robustness in Machine Learning} The robustness literature in machine learning has primarily focused on robustness to adversarially perturbed inputs, which add small amounts of carefully selected noise to fool classifiers \cite{Goodfellow15}. When applied to the text setting, such adversarial examples can be generated by a variety of editing methods, including through translation \cite{SEAR}, attributions \cite{Mud18}, and autoencoders  \cite{GNAE} \cite{Hu17}. Adversarial examples are closely related to counterfactual examples: \citeauthor{Wachter17} characterize counterfactuals as adversarial examples that perturb inputs in human-interpretable and possibly problematic ways. As such, the counterfactual examples presented in this work can be viewed as a specific subset of adversarial examples. The robustness literature has attempted to address adversarial examples using a variety of techniques, such as adversarial training \cite{Madry17,Goodfellow15} and adversarial logit pairing \cite{ALP}. 

Several papers draw connections between fairness, text generation, and robustness. \citeauthor{TEXT_ROBUST} consider robustness in text with respect to counfounding variables such as the author's gender, and learn robust models by training using an additional attribute for the latent confound, and averaging over all values of the latent variable at inference time. \citeauthor{Madaan18} attempt to edit text to remove gender bias or edit gender representations, leveraging analogies in word vector differences to make substitutions for words that may implicitly encode biases about gender. 

\section{Problem Definition}
Given text input $x \in X$, where $x$ is a sequence $[x_1, ..., x_n]$ of tokens, our task is to predict an outcome $y$. We consider a classifier $f$ parameterized by $\theta$ that produces a prediction $\hat{y} = f_\theta(x)$, where we seek to minimize some notion of error between $y$ and $\hat{y}$. For notational simplicity, we restrict the following definitions to a single binary class, but they can be easily generalized to multi-class classification problems. The classifier $f$ can be an arbitrary neural network.

We wish to maximize the model's performance while maintaining counterfactual fairness with respect to sensitive attributes, such as identity groups. Counterfactual fairness is measured using counterfactual examples that perturb the sensitive attribute referenced in the example at hand.
Let $\Phi(x)$ denote the set of counterfactual examples associated with an example $x$. Counterfactual fairness requires that the predictions of a model for all counterfactuals are within a specified error. 

\begin{definition} \textit{Counterfactual fairness}. A classifier $f$ is counterfactually fair with respect to a counterfactual generation function $\Phi$ and some error rate $\epsilon$ if
$$|f(x) - f(x')| \leq \epsilon \quad \forall x \in X, x' \in \Phi(x)$$
\end{definition}

\subsection{Counterfactual Token Fairness (CTF)} \label{ctf}
We consider a narrow class of counterfactuals that involves substituting identity tokens in the input, for instance, substituting ``gay'' with ``straight'' in the input ``Some people are gay.'' 
We assume a set of identity tokens, $\mathcal{A}$, for which we seek to be counterfactually fair. Consider a pair of tokens $a, a' \in \mathcal{A}$. The associated counterfactual example generation function $\Phi_{a, a'}$ is defined by substituting all occurrences of $a$ in $x$ with $a'$ and vice versa. If neither identity token is present in the input $x$, then $\Phi_{a, a'}(x) = \emptyset$. 
We generalize this definition to a counterfactual generation function over $\mathcal{A}$ that generates all counterfactual examples based on pairs of substitutions:
$$\Phi_{\mathcal{A}}(x) = \bigcup_{a \neq a' \in \mathcal{A}} \Phi_{a, a'}(x)$$

\begin{definition}
A classifier satisfies \emph{counterfactual token fairness} with respect to a set of identity tokens $\mathcal{A}$ if it satisfies counterfactual fairness with respect to the counterfactual generation function $\Phi_{\mathcal{A}}$ and error rate $\epsilon$.
\end{definition}

Although content about sensitive groups may be captured by complex semantics, this metric will surface a subset of problematic issues related to more general counterfactual fairness. This a first step, and surfaces additional concerns for fairness beyond those of group fairness.

\subsection{Asymmetric Counterfactuals}

So far we have assumed that all counterfactuals with respect to identity tokens should have the same prediction. This assumption is not valid in cases where the sensitive attribute indeed affects the prediction. 
For instance, consider a model predicting toxicity of text, and the counterfactual pair ``That's so gay'' and ``That's so straight.'' The first example is arguably more likely to be considered toxic than the second, as ``gay'' is often used as an insult in Internet forums, while ``straight'' is not. Other examples include stereotyping, where one group is more vulnerable than another. Requiring equal predictions across such cases can inadvertently harm the more vulnerable group.

Fairness must be required only among counterfactuals that stipulate symmetric predictions. This restriction can be accommodated in our framework by restricting the counterfactual generation function $\Phi(x)$ to exclude any counterfactuals for the example $x$ that may have asymmetric labels. In general, the degree and direction of the asymmetry between counterfactuals varies based on the task, and the cultural sensitivities of the consumers of the task. This makes it difficult to define a perfect counterfactual generation function. In Experiments, we propose a 
heuristic for avoiding asymmetric counterfactuals for a model predicting toxicity of text.

\subsection{Relationship to Group Fairness}\label{relationship}
Counterfactual fairness is complementary to the group fairness notion of \emph{equality of odds}~\cite{Hardt16}, which demands equality of true positive rates and true negative rates for different values of the sensitive attribute. A text classifier may satisfy one while completely failing the other.
Consider the case when two sensitive attributes $a$ and $a'$ only appear in disjoint sets of contexts $X_a$ and $X_{a'}$, respectively. A model can satisfy equality of odds by always predicting correctly on the contexts in which $a, a'$ appear in the data, but never in the counterfactual contexts that do not exist in the data. Conversely, the model could predict identical output for all counterfactual pairs while predicting correctly only on $X_a$ and not $X_a'$.

\section{Methods}

We propose three methods to improve counterfactual fairness: blindness, counterfactual augmentation, and counterfactual logit pairing. Both methods assume access to a list of identity tokens for which they seek to be fair.

\subsection{Blindness}
Blindness substitutes all identity tokens with a special \identity token, which allows the predictor to know that an identity term is present, but not which identity. This is similar to standard NLP methods such as replacing large numbers with a generic \texttt{NUMBER}\xspace.
While this approach guarantees counterfactual token fairness, it has a number of downsides.
First, it does not have the ability  to differentiate identity terms, and so necessarily equates asymmetric counterfactuals. Second,
it cannot handle complex counterfactuals that involve more than single token substitutions, e.g. ``Christians go to church.'' and ``Jews go to temple.''
Finally, the model may still discriminate using other signals that are associated with the identity term \cite{Dwork11}.

\subsection{Counterfactual Augmentation}
Instead of blinding the model to identity terms, counterfactual augmentation involves augmenting the model's training set with generated counterfactual examples. The additional examples are meant to guide the model to become invariant to perturbing identity terms. This is a standard technique in computer vision for making the model invariant to object location, image orientation, etc. The counterfactual examples are assigned the same label as the original example. 

\subsection{Counterfactual Logit Pairing (CLP)}
Counterfactual logit pairing (CLP) encourages the model to be robust to identity by adding a robustness term to the training loss. The robustness term is given by logit pairing \cite{ALP}, which penalizes the norm of the difference in logits for pairs of training examples and their counterfactuals.
Specifically, suppose the classifier $f(x) = \sigma(g(x))$, where $g(x)$ produces a logit and $\sigma(\cdot)$ is the sigmoid function. 
The additional loss is the average absolute difference in logits between the inputs and their counterfactuals: 
\[\sum_{x \in X}{\mathop{\mathbb{E}}_{x' \sim \text{Unif}[\Phi(x)]}|g(x) - g(x')|}\]
For computational tractability, during training, we randomly sample a single counterfactual example for each input. Taking $J$ as the original loss function, the overall objective is:
\[\sum_{x \in X}{J(f(x), y)} +  \lambda\sum_{x \in X}{\mathop{\mathbb{E}}_{x' \sim \text{Unif}[\Phi(x)]}|g(x) - g(x')|}\]
Similar to counterfactual augmentation, CLP can use any counterfactual generation function. For example, a restricted counterfactual generation function could be used to avoid enforcing equality over asymmetric counterfactuals. Moreover, the method also applies if more sophisticated counterfactuals are generated.

In contrast to counterfactual augmentation, the robustness term in the CLP loss explicitly guides the model to satisfy two desirable properties: (1) ensuring a model produces similar outputs on counterfactual pairs and (2) learning models that generalize well to different identities. 
Moreover, the parameter $\lambda$ can be tuned to achieve varying degrees of counterfactual fairness.

\section{Experiments} \label{experiments}
\paragraph{Dataset} We evaluate our methods on the task of predicting toxicity. For the task, a toxic comment is defined as a ``rude, disrespectful, or unreasonable comment that is likely to make you leave a discussion.'' \cite{Dixon18}. We use a public Kaggle dataset of 160K Wikipedia comments, each labeled by human raters as toxic or non-toxic\footnote{https://www.kaggle.com/c/jigsaw-toxic-comment-classification-challenge}, randomly split into train and dev sets. We evaluate AUC of the primary task on the public test set. We evaluate counterfactual token fairness and group fairness on a private dataset of comments from another internet forum. This dataset, henceforth called the ``evaluation dataset,'' has a higher occurrence of identity terms, and therefore leads to a more meaningful fairness evaluation.

\paragraph{Setup} We evaluate our methods for counterfactual token fairness on the set of 50 identity terms used by \citeauthor{Dixon18}. Out of these, 47 are single tokens and 3 are bigrams. We randomly partition the terms into a training set of 35 and a hold-out set of 12 to evaluate generalization. We also include the three bigrams in evaluation, because they reflect scenarios that blindness cannot address during training.\footnote{Only single tokens in the input are substituted with bigrams during evaluation.} 

All of the models are CNNs trained with cross entropy loss against the binary toxicity label. All hyperparameters except for the fairness regularizer $\lambda$ for CLP were fixed for all runs of all models. Models were trained for five epochs, and the best model on the dev set was taken. Each model was trained ten times, and the average of the runs is reported. Blindness, Counterfactual augmentation, and CLP models (for different values of $\lambda$)  were evaluated and compared to a baseline model. 

For CLP training, we define a different counterfactual example generation function than the one used for evaluation. The evaluation counterfactuals only apply substitutions to a pair of identity tokens, whereas during training, each sensitive identity token in an input is randomly substituted with another identity token. 

\paragraph{Handling Asymmetric Counterfactuals}
We hypothesize that asymmetric counterfactuals are less likely to arise for ground truth non-toxic comments than toxic comments. This for two reasons. 
Asymmetric counterfactuals arise when stereotyping / attacking a vulnerable group occurs for some identity substitution, and no other toxicity signals are present.
In such cases, most identity substitutions will be nontoxic, and only the one attacking the vulnerable group(s) will be toxic. So if the ground truth example is nontoxic, counterfactual fairness can still be required over most identity substitutions, whereas if the ground truth example is toxic, equal prediction should not be required over most counterfactuals.
Second, the stereotyping comments are more likely to occur in a toxic comment attacking the stereotyped group than in a nontoxic comment referencing some other identity group. 
For these reasons, we evaluate counterfactual token fairness over ground truth nontoxic comments separately from ground truth toxic comments, and focus our analysis on nontoxic comments.
We also consider applying the CLP loss only to nontoxic comments during training, to avoid enforcing equality of logits for potentially asymmetric counterfactuals. We distinguish this variant as CLP\_nontoxic.

Separately, we also evaluate CTF on simple synthetic inputs where all information about toxicity is encoded in the context, and all counterfactuals are symmetric by design. Specifically, we use a dataset of synthetically generated sentences based on templates such as ``\texttt{NAME}\xspace is a \texttt{ADJECTIVE}.''\footnote{This is the updated open sourced version of the synthetic test set presented in \cite{Dixon18}.}

\paragraph{Metrics}
We measure the \emph{counterfactual token fairness gap} with respect to a given counterfactual generation function. For a single example, this is the average gap in prediction over all of the counterfactual pairs for that example: 
\[\text{CF GAP}_{\Phi}(x) = {\mathop{\mathbb{E}}_{x' \sim \text{Unif}[\Phi(x)]}|f(x) - f(x')|}\]
Over an entire dataset, the gap is the average over all examples that have valid counterfactuals. In this study, we measure the CTF GAP for the counterfactual generation function $\Phi_{\mathcal{A}}$, which substitutes all pairs of identity tokens.
Because substitution-based counterfactuals over short inputs are more likely to be logical, we evaluate the CTF gaps for inputs of at most ten tokens in length. 
In addition, since asymmetric counterfactuals are likely more common for toxic comments, we evaluate CTF gaps over nontoxic and toxic comments separately. 

We also measure group fairness to ensure that optimizing for counterfactual fairness has no perverse impact on it. Following the group fairness notion of equality of odds~\cite{Hardt16}, we measure the true positive rates (TPR) and true negatives rates (TNR) of examples referencing different identity groups. We assume an example references a specific identity group based on the presence of the associated token. 
Equality of odds requires equal TPR and TNR across identities, so we evaluate overall TPR and TNR gaps. The gap for a pair of identity terms is computed as the absolute value of the difference in rates for the two identity terms. The overall TPR or TNR gap is the average over all pairs of identity terms.

\subsection{Results}

\begin{table}[]
\begin{tabular}{|l|l|l|l|}
\hline
Model                      & Eval NT & Synth NT & Synth Tox \\ \hline
Baseline                   & 0.140             & 0.180              & 0.061           \\ \hline
Blind                      & 0.000             & 0.000              & 0.000           \\ \hline
CF Aug            & 0.127             & 0.226              & 0.022           \\ \hline
CLP\_nontox, $\lambda=1$ & 0.012             & 0.015              & 0.007           \\ \hline
CLP,  $\lambda=0.05$       & 0.071             & 0.082              & 0.024           \\ \hline
CLP,  $\lambda=1$          & 0.007             & 0.015              & 0.007           \\ \hline
CLP,  $\lambda=5$          & 0.002             & 0.004              & 0.004           \\ \hline
\end{tabular}
\caption{Conterfactual token fairness gaps for non-toxic examples from evaluation set and all examples from a synthetic test set.
All gaps are measure w.r.t. 35 training terms. Smaller gaps are better. 
}
\label{tab:ctf}

\end{table}

\paragraph{Counterfactual Token Fairness}
Table~\ref{tab:ctf} reports CTF gaps for non-toxic examples from the evaluation dataset, and all examples from the synthetic dataset. The gaps are computed for the 35 training terms (discussed in Setup). As discussed earlier, both these sets of examples are unlikely to have asymmetric counterfactuals. The baseline model has a large CTF gap on both sets of examples.
Blindness achieves a zero gap by design. CLP with a fairness regularization coefficient ($\lambda$) of at least 1 also attains a near zero gap. Counterfactual augmentation decreases the CTF gap (relative to the baseline) on non-toxic examples from the evaluation dataset, but does not obtain a zero gap.
It is worth noting that the models were not trained on the synthetic dataset, but we still find a reduction in counterfactual fairness gaps on it. 

\begin{table}[]
\centering
\begin{tabular}{l|l|}
\cline{2-2}
                                               &  CTF Gap: held-out terms \\ \hline
\multicolumn{1}{|l|}{Baseline}                 & 0.091       \\ \hline
\multicolumn{1}{|l|}{Blind}                    & 0.090       \\ \hline
\multicolumn{1}{|l|}{CF Aug}                   & 0.087       \\ \hline
\multicolumn{1}{|l|}{CLP\_nontox, $\lambda=1$} & 0.095       \\ \hline
\multicolumn{1}{|l|}{CLP,  $\lambda=0.05$}     & 0.078       \\ \hline
\multicolumn{1}{|l|}{CLP,  $\lambda=1$}        & 0.084       \\ \hline
\multicolumn{1}{|l|}{CLP,  $\lambda=5$}        & 0.076       \\ \hline
\end{tabular}
\caption{CTF gaps on held out identity terms for non-toxic examples from the evaluation set.}\label{tab:ctf_generalization}
\end{table}

Table~\ref{tab:ctf_generalization} reports CTF gaps on the hold-out terms for non-toxic examples from the evaluation dataset.  We say that a model's CTF gap generalizes to hold-out terms if its gap is less than the baseline model's gap (0.091). Among the models compared, CLP with $\lambda=5$ generalizes the best, though the gaps are much larger that those on the training terms. Blindness does not appear to provide generalization benefits. Thus, it may not be a favorable method in settings where we expect examples with identity terms outside the set of training terms.

To evaluate the impact on cases with asymmetric counterfactuals, we also measured the CTF gap for toxic examples from the evaluation dataset over the 35 training terms; see Table~\ref{tab:ctf_toxic} in the appendix. The baseline model has a gap of 0.241, and as expected, blindness has a gap of zero. All CLP models with $\lambda \geq 1$ achieve a CTF gap of less than 0.03, which unfortunately means that they end up equating predictions for asymmetric counterfactuals. This includes CLP\_nontoxic, which was trained using counterfactuals from non-toxic examples only. Going forward, we do not evaluate CLP\_nontoxic as it is not better than the regular CLP models.

\paragraph {Overall Performance}
We evaluate the overall classifier performance using AUC of the ROC curve. Remarkably, all methods show consistent AUC on the test set, ranging between 0.962-0.964. 

\begin{figure}[h]

    \centering  
    \includegraphics[width=0.47\textwidth]{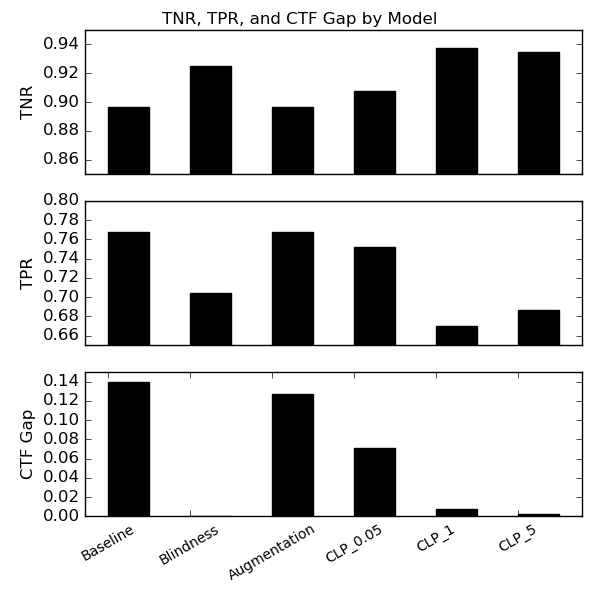}
    \caption{Plot of the average CTF gap along with the TPR and TNR over examples that contain identity terms.}
    \label{fig:tprtnr}
    \end{figure}

Figure \ref{fig:tprtnr} compares the true positive rate (TPR) and true negative rate (TNR) of various models, where the threshold for a toxic classification is set to 0.5. TPR and TNR are measured only over examples that contain an identity term from the set of training terms. We find that methods that reduce the CTF gap perform better at identifying nontoxic comments (true negatives) and worse at identifying toxic comments (true positives). We discuss this tension between improving the CTF gap and TPR in Error Analysis. 

\paragraph{Group Fairness}
We additionally evaluate the group fairness metrics, TPR and TNR gaps for equality of odds (see Table~\ref{tab:eo}). Counterfactual augmentation and CLP with $\lambda=0.05$ have better TPR and TNR gaps than the baseline and are able to reduce CTF gaps. CLP with $\lambda \geq 1$ has a more extreme tradeoff, harming the TPR gap while substantially improving the TNR gap. Practitioners may choose different tradeoffs of CTF gap, TPR, and TNR depending on the relative prioritization of these metrics for a given task.

\begin{table}[]
\centering
\begin{tabular}{l|l|l|}
\cline{2-3}
                                              & TNR Gap & TPR Gap \\ \hline
\multicolumn{1}{|l|}{Baseline}                & 0.084      & 0.082      \\ \hline
\multicolumn{1}{|l|}{Blindness}               & 0.039      & 0.114      \\ \hline
\multicolumn{1}{|l|}{Augmentation}            & 0.065      & 0.083      \\ \hline
\multicolumn{1}{|l|}{CLP all, $\lambda=0.05$} & 0.058      & 0.078      \\ \hline
\multicolumn{1}{|l|}{CLP all, $\lambda=1$}    & 0.039      & 0.104      \\ \hline
\multicolumn{1}{|l|}{CLP all, $\lambda=5$}    & 0.041      & 0.112      \\ \hline
\end{tabular}
\caption{TNR and TPR gaps for different models. Lower is better.}\label{tab:eo}
\end{table}

\subsection{Error analysis}\label{sec:error}

We examine the trade-off between CTF gap and TPR. We consider the CLP, $\lambda=5$ model which attains a near zero CTF gap and compare its predictions on toxic comments to those of the baseline. Among examples with identity terms in the test set, there are 83 cases where an example was correctly classified by the baseline and incorrectly classified by the CLP model. Of these, 27 were labeled by an author as having an asymmetric counterfactual. There were 20 cases where the CLP model predicted correctly compared to the baseline, of which none had asymmetric counterfactuals. This tells us that a large chunk of the TPR loss (relative to the baseline) is over toxic examples with asymmetric counterfactuals.
This is expected as examples with asymmetric counterfactuals are toxic because of the presence of a specific identity term, and a model trained to disregard identity terms will be less likely to predict correctly on such examples. 

As a means of investigating what the CLP model has learned, we examine its token embeddings after convergence. By the end of training with $\lambda=5$, the average cosine similarity between pairs of identity tokens is 0.87, whereas the baseline has an average cosine similarity of 0.25. Although this is similar to blindness, the CLP model learns a different toxicity association with identity tokens. The average toxicity prediction on a single identity token for the CLP model is 0.12, while the toxicity of the \identity token in the blindness model is 0.54.

Similarly, CLP\_nontoxic with $\lambda=5$ has a average cosine similarity of 0.81. This embedding convergence, despite CLP being applied only to nontoxic comments, is the reason why the model achieves a low CTF gap on toxic comments, including those with asymmetric counterfactuals. Methods to enforce equal prediction on some subset of counterfactuals but not others should be further investigated.

We also qualitatively evaluate the strength of each model's association with various identity tokens. Table \ref{tab:qea} in the appendix lists various examples, and the associated toxicity scores from each model. In contrast to the baseline, all three models associate a much smaller amount of toxicity signal with the identity tokens. For instance, unlike the baseline, the other models no longer associate a substantial amount of toxicity with clearly nontoxic statements such as ``Some people are gay.'' Notably, the toxicity of the statement ``Some people are straight'' goes up. The negative effect on this pair is more pronounced for blindness than it is for CLP.

\section{Conclusions and Future Work} \label{disc}
We make progress towards counterfactual fairness in text classification. We propose a specific form of counterfactual fairness, called \emph{counterfactual token fairness} (CTF), that requires a model to be robust to different identity tokens present in the input. We show that text classification models with good overall performance fare poorly on this metric. 
We approach counterfactual token fairness from a robustness perspective, and offer a procedure, \emph{counterfactual logit pairing}, for optimizing the counterfactual token fairness metric during model training. 
We find that this approach performs as well as blindness to identity tokens, but also generalizes better to hold-out tokens. 
These results do not come at the expense of overall classifier accuracy, and have varying tradeoffs between false positives and false negatives.

Going forward, better heuristics must be designed for identifying cases with asymmetric counterfactuals. Excluding toxic comments covers many but not all asymmetric examples. For example, ground truth nontoxic examples referencing ``black power'' are more likely to become toxic as they reference ``white power.'' In other text classification tasks such as sentiment classification, asymmetric counterfactuals will arise but not necessarily with the same clear split by label.

A next step would be to improve counterfactual generation by addressing issues of polysemy of identity terms (which can result in illogical substitutions), asymmetric counterfactuals, and multiple references to an identity group. One possible method is to use analogies in word vectors to change multiple tokens used for the same identity group \cite{Madaan18}. Another approach is defining a generative model over text, as in \cite{Hu17}, that can modify certain attributes of the text while holding others constant and preserving semantics. 
One could also use criteria for selecting semantically equivalent adversarial examples as in \cite{SEAR}, to evaluate whether counterfactual examples are logical. Optimizing for general counterfactual fairness will test many of the unique advantages of counterfactual logit pairing. 

\subsection{Acknowledgements}
We would like to thank Lucy Wasserman, Allison Woodruff, Yoni
Halpern, Andrew Smart, Tulsee Doshi, Jilin Chen, Alexander D’Amour,
Raz Mathias, and Jon Bischof for their feedback leading up to this
paper.

\bibliographystyle{aaai}
\bibliography{sample-bibliography}

\clearpage
\onecolumn
\appendix
\section{Appendix}

\begin{table}[h]
\centering
\begin{tabular}{l|l|l|}
\cline{2-3}
                                                 & Train Terms & Held-out Terms \\ \hline
\multicolumn{1}{|l|}{Baseline}                   & 0.241       & 0.071          \\ \hline
\multicolumn{1}{|l|}{Blind}                      & 0.000       & 0.062          \\ \hline
\multicolumn{1}{|l|}{CF Augmentation}            & 0.155       & 0.057          \\ \hline
\multicolumn{1}{|l|}{CLP\_nontoxic, $\lambda=1$} & 0.029       & 0.068          \\ \hline
\multicolumn{1}{|l|}{CLP,  $\lambda=0.05$}       & 0.165       & 0.057          \\ \hline
\multicolumn{1}{|l|}{CLP,  $\lambda=1$}          & 0.010       & 0.058          \\ \hline
\multicolumn{1}{|l|}{CLP,  $\lambda=5$}          & 0.004       & 0.051          \\ \hline
\end{tabular}
\caption{CTF gaps on toxic examples from the evaluation set, for both training terms and held-out terms.}\label{tab:ctf_toxic}

\end{table}

\begin{table*}[h]
\centering
\begin{tabular}{l|l|l|l|l|}
\cline{2-5}
                                                & Baseline & Blindness & CF Augmentation & CLP , $\lambda=5$ \\ \hline
\multicolumn{1}{|l|}{Some people are gay}       & 0.98     & 0.61      & 0.82                        & 0.14              \\ \hline
\multicolumn{1}{|l|}{Some people are straight}  & 0.02     & 0.61      & 0.11                        & 0.14              \\ \hline
\multicolumn{1}{|l|}{Some people are Jewish}    & 0.28     & 0.61      & 0.17                        & 0.13              \\ \hline
\multicolumn{1}{|l|}{Some people are Muslim}    & 0.46     & 0.61      & 0.24                        & 0.14              \\ \hline
\multicolumn{1}{|l|}{Some people are Christian} & 0.04     & 0.16      & 0.02                        & 0.14              \\ \hline
\end{tabular}

\caption{Counterfactuals and toxicity scores of different models. The tokens ``gay,'' ``straight,'' ``jewish,'' and ``muslim'' are used during training, and ``christian'' was held-out.}
\label{tab:qea}
\end{table*}

\end{document}